\pdfoutput=1

\documentclass[11pt]{article}

\usepackage[final]{coling}

\usepackage{times}
\usepackage{latexsym}

\usepackage[T1]{fontenc}

\usepackage[utf8]{inputenc}

\usepackage{microtype}

\usepackage{inconsolata}

\usepackage{graphicx}

\usepackage{booktabs}
\usepackage{amsmath}
\usepackage{amssymb}
\usepackage{pifont}
\usepackage{multirow}
\usepackage{url}
\usepackage[capitalise, noabbrev]{cleveref}
\usepackage[shortcuts, acronym, hyperfirst = false]{glossaries}
\usepackage[symbol]{footmisc}

\DeclareMathOperator*{\argmax}{arg\,max}

\DeclareMathOperator*{\onehot}{one\_hot}
\DeclareMathOperator*{\softmax}{Softmax}

\newcommand{\gqa}{GQA}

\newcommand{\chatgpt}{GPT4}
\newcommand{\gemini}{Gemini}
\newcommand{\llama}{Llama}
\newcommand{\llava}{LLaVA}

\newcommand{\imle}{\textsc{Imle}}
\newcommand{\aimle}{\textsc{Aimle}}
\newcommand{\simple}{\textsc{Simple}}
\newcommand{\gumbelst}{\textsc{Gumbel SoftSub-ST}}

\title{Discrete Subgraph Sampling for Interpretable Graph based Visual Question Answering}

\author{Pascal Tilli \\
  University of Stuttgart \\ Stuttgart, Germany \\
  \texttt{pascal.tilli@ims.uni-stuttgart.de} \\\And
  Ngoc Thang Vu \\
  University of Stuttgart \\ Stuttgart, Germany \\
  \texttt{thang.vu@ims.uni-stuttgart.de} \\}

\begin{document}

\maketitle

\newacronym{gnn}{GNN}{Graph Neural Network}
\newacronym{gnns}{GNNs}{Graph Neural Networks}
\newacronym{sgg}{SGG}{Scene Graph Generation}
\newacronym{vqa}{VQA}{Visual Question Answering}
\newacronym{okvqa}{VQA}{Visual Question Answering}
\newacronym{gvqa}{GVQA}{Graph-based Visual Question Answering}
\newacronym{qa}{QA}{Question Answering}
\newacronym{dl}{DL}{Deep Learning}
\newacronym{ml}{ML}{Machine Learning}
\newacronym{nlp}{NLP}{Natural Language Processing}
\newacronym{cv}{CV}{Computer Vision}
\newacronym{gat}{GAT}{Graph Attention Network}
\newacronym{gats}{GATs}{Graph Attention Networks}
\newacronym{xai}{XAI}{eXplainable Artificial Intelligence}
\newacronym{ai}{AI}{Artificial Intelligence}
\newacronym{imle}{\textsc{IMLE}}{Implicit Maximum Likelihood Estimation}
\newacronym{pid}{PID}{perturbation-based implicit differentiation}
\newacronym{aimle}{\textsc{AIMLE}}{Adaptive Implicit Maximum Likelihood Estimation}
\newacronym{nsm}{NSM}{Neural State Machine}
\newacronym{rl}{RL}{Reinforcement Learning}
\newacronym{mgat}{M-GAT}{Masking Graph Attention Network}
\newacronym{subgat}{Sub-GAT}{Subgraph Graph Attention Network}
\newacronym{kergnns}{KerGNNs}{Kernel Graph Neural Networks}
\newacronym{wl}{WL}{Weisfeiler-Lehman}
\newacronym{cnn}{CNN}{Convolutional Neural Network}
\newacronym{cnns}{CNNs}{Convolutional Neural Networks}
\newacronym{mlp}{MLP}{Multi-Layer Perceptron}
\newacronym{gnnex}{GNNExplainer}{GNNExplainer}
\newacronym{pyg}{PyG}{PyTorch Geometric}
\newacronym{atcoo}{AT-COO}{Answer Token Co-occurrence}
\newacronym{qtcoo}{QT-COO}{Question Token Co-occurrence}
\newacronym{map}{MAP}{Maximum A-Posteriori}
\newacronym{ste}{STE}{Straight-Through estimator}
\newacronym{clip}{CLIP}{Contrastive Language-Image Pre-Training}

\begin{abstract}
    Explainable artificial intelligence (XAI) aims to make machine learning models more transparent.
    While many approaches focus on generating explanations post-hoc, interpretable approaches, which
    generate the explanations intrinsically alongside the predictions, are relatively rare.
    In this work, we integrate different discrete subset sampling methods into a
    graph-based visual question answering system to compare their effectiveness in generating
    interpretable explanatory subgraphs intrinsically.
    We evaluate the methods on the \gqa~dataset and show that the integrated methods effectively
    mitigate the performance trade-off between interpretability and answer accuracy, while also achieving
    strong co-occurrences between answer and question tokens.
    Furthermore, we conduct a human evaluation to assess the interpretability of the generated
    subgraphs using a comparative setting with the extended Bradley-Terry model, showing that the 
    answer and question token co-occurrence metrics strongly correlate with human preferences.
    Our source code is publicly available\footnote[2]{\url{ https://github.com/DigitalPhonetics/Intrinsic-Subgraph-Generation-for-VQA}}.
\end{abstract}

\section{Introduction}
With the rise of foundational models such as \llava~\cite{llava,improvedllava,llavanext}, 
\chatgpt~\cite{chatgpt}, \gemini~\cite{gemini}, or \llama~\cite{llama}, the need for interpretable 
and explainable \gls*{ml} systems has become increasingly apparent, which is reflected by the increasing
number of publications in \gls*{xai}~\cite{xai_survey}.

In this work, we focus on interpretable \gls*{gvqa}, which involves answering questions 
about images.
Specifically, we use scene graph representations of the visual input~\cite{gqa} instead of the 
raw images, as proposed in the original versions of \gls*{vqa}~\cite{vqa, vqa_v2}.
Scene graphs have been successfully applied to \gls*{vqa} 
tasks~\cite{sg_vqa_survey, understanding_sg_vqa, graphvqa, vqa_gnn} and have shown the potential to 
generate subgraphs as explanations intrinsically~\cite{tilli}.

The \gls*{gvqa} approach by \citet{tilli} discretely samples a subgraph based on 
\gls*{imle}~\cite{imle} and compares the results to post-hoc explainability methods with quantitative 
metrics, i.e.,~\gls*{atcoo} and \gls*{qtcoo}, as well as qualitatively through human evaluation.
This raises the question of how different intrinsic subgraph sampling methods compare to each other 
, and how the proposed metrics generalize in contexts without post-hoc methods.

Sampling a subset from a complex discrete distribution is ubiquitous in \gls*{ml} and 
has many applications.
However, while these methods~\cite{gumbel-softmax-1,gumbel-softmax-2, imle, aimle, simple} have been 
proposed and applied in various fields, they have never been explored in the multi-modal context of \gls*{gvqa}.
Hence, we pose the following research questions:
\begin{description}
    \item[\textbf{RQ1}] What is the effect of different discrete subgraph sampling methods on 
    question-answering performance, as well as answer and question token co-occurrences?
    \item[\textbf{RQ2}] Do the answer and question token co-occurrence metrics generalize in a human 
    evaluation when different discrete subgraph sampling methods are used?
\end{description}

To address these research questions, we propose integrating \aimle, \simple, and \gumbelst~into a 
\gls*{gvqa} system to compare the effectiveness with \imle, as introduced by \citet{tilli}.
We evaluate the models' performances in terms of answer accuracy, \gls*{atcoo}, and \gls*{qtcoo}.
To verify the effectiveness of the \gls*{atcoo} and \gls*{qtcoo} metrics when comparing different 
discrete subgraph sampling methods, we conduct a human evaluation to assess the 
interpretability of the generated subgraphs in a comparative setting with human evaluators.

Our contributions are as follows: 
\textbf{(1)} We demonstrate that these methods effectively mitigate the performance trade-off between 
interpretability and answer accuracy, while also achieving strong answer and question token co-occurrences.
\textbf{(2)} We show that the answer and question token co-occurrence metrics strongly 
correlate with human preferences, highlighting the effectiveness of the metrics in capturing the 
relevant aspects of the question answering process.

\section{Methods}

\subsection{Top-\emph{k} Subgraph Sampling}
\paragraph{Interpretability}
Our model generates a subgraph most relevant to a given question as an explanation alongside 
the answer prediction.
To achieve this, we employ subgraph sampling during both training and inference, a process complicated 
by the inherent non-differentiability of sampling from discrete distributions.

\paragraph{Top-\emph{k} Sampling}
We integrate a top-$k$ combinatorial solver into our system, enabling it to identify the most 
relevant nodes (subgraph) for a question.
This is achieved by deriving a discrete solution from $z \leftarrow \texttt{topk}(\theta)$, 
$z \in \{0,1\}^n$, which is equivalent to computing the \gls*{map} state, 
i.e.,\ the most probable state
\begin{equation}
    \texttt{MAP}(z) \equiv \argmax_{z} p_\theta(z)
\end{equation}\label{eq:map}
of an exponential family distribution, where $z \sim p_\theta(z)$.
$\theta$ refers to the prior scores (computed by our model) that parameterize the distribution 
$p_\theta(z)$
In our case, this is a conditional distribution based on a top-$k$ constraint, 
$z \sim p_\theta(z|\sum_{i=1}z_i = k)$.

\paragraph{Discrete Gradient Computation}
Computing the gradient $\nabla_\theta L(f(z), \hat{y})$ poses significant challenges due to the 
discrete nature of $z$.
Here, $L$ denotes our loss function, $f(z)$ is the model's output based on the subgraph $z$, and 
$\hat{y}$ represents the ground-truth labels.
We incorporate recent methods to approximate $\nabla_\theta L$, which we will briefly outline below.
These methods have strong theoretical foundations and are practically efficient.
We select them because they provide stable, low-variance estimates of the gradients and integrate seamlessly into end-to-end optimization frameworks.

\subsubsection{\gumbelst}
The \textsc{Gumbel-Softmax} leverages the \textsc{Gumbel-Max} trick~\cite{perturb_map, gumbel-max} to sample from a discrete probability distribution by perturbing the logits with standard Gumbel noise. 
Specifically, given logits $\theta$, we draw $g_i \sim \text{Gumbel}(0,1)$ for each $i$ and select \begin{equation} z = \onehot(\argmax_{i \in {1,\ldots,n}}(\theta_i + g_i)) \sim p_\theta, \end{equation} where $\onehot$ converts the input into a one-hot encoded vector, i.e. a binary vector.
Since $\argmax$ is not differentiable, the \textsc{Gumbel-Softmax} trick relaxes it to a $\softmax$ operation with $y = \softmax(\theta_i + g_i)$, using $y$ as a continuous proxy for the discrete $z$. 
Building on this, the \gumbelst~\cite{softsub-st} method extends the \textsc{Gumbel-Softmax} trick~\cite{gumbel-softmax-1, gumbel-softmax-2} to enable sampling of relaxed top-$k$ subsets, maintaining differentiability and allowing for backpropagation through the sampling step.

\subsubsection{\textsc{Imle} and \textsc{Aimle}}

\imle~\cite{imle} with \gls*{pid} target distributions generalizes the \gls*{ste} to more complex 
distributions.
Instead of directly using the gradients $\nabla_z L$ for backpropagation, \imle~leverages them to 
construct a target distribution $q$.
This defines an implicit maximum likelihood objective, whose gradient estimator propagates the 
supervisory signal upstream.
More formally, the gradient is approximated as
\begin{equation}
    \nabla_{\theta} L \approx \frac{1}{\lambda} [\texttt{MAP}(\theta + \epsilon) - \texttt{MAP}(\theta' + \epsilon)]
\end{equation}
where $\epsilon \sim p(\epsilon)$ is drawn from a Gumbel(0,1) distribution and 
$\theta' = \theta - \lambda \nabla_z L(f_u(z), \hat{y})$.
To efficiently sample from $p_\theta(z)$, Perturb-and-MAP~\cite{perturb_map} is applied with 
$z = \texttt{MAP}(\theta + \epsilon)$.
A key idea is the construction of a target distribution $q$ based on the prior distribution $p$ 
using \gls*{pid}
\begin{equation}
    q(z, \theta') = p(z, \theta - \lambda \nabla_z L(f_u(z), \hat{y}))
\end{equation}

\gls*{aimle}~\cite{aimle} extends~\imle~by adaptively changing $\lambda$ until the gradient 
estimates meet a desired sparsity criterion.
This adaptation is guided by an update rule where $\lambda$ is adjusted according to the exponential 
moving average of the gradient's $L_0$-norm, ensuring that the gradient estimates achieve a desired 
level of non-zero elements.

\subsubsection{\simple}

\citet{simple} address the challenge of gradient estimation for sampling from a $k$-subset 
distribution, which is computationally intractable due to the combinatorial nature of the problem. 
The gradient depends on the marginals of the distribution $\mu(\theta)$, which are the partial 
derivatives of the log-probability of the $k$-subset constraint.
They proposed \simple, a method that efficiently computes these marginals to approximate the gradient 
while reducing the computational complexity compared to exact methods.
The gradient is approximated as
\begin{equation}
    \nabla_{\theta} L \approx \partial_{\theta} \mu(\theta) \nabla_{z} L
\end{equation}

\subsection{Graph-based VQA System}
\begin{figure}[htb!]
    \centering
    \resizebox{\linewidth}{!}{
    \includegraphics{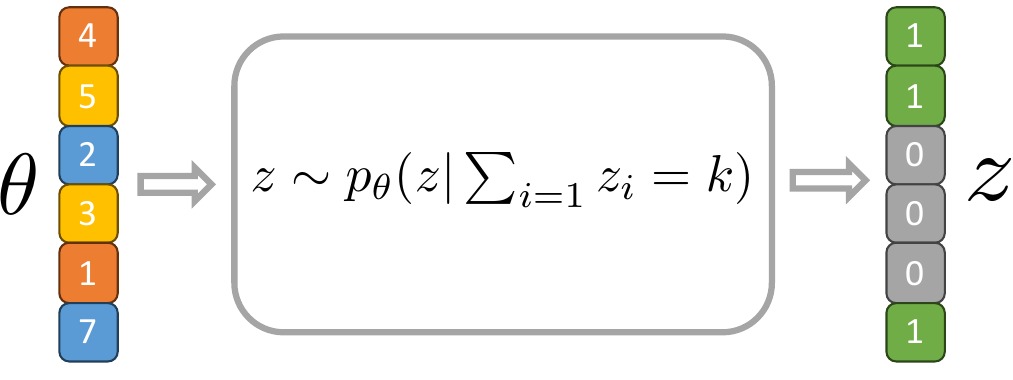}
    }
    \caption{The prior scores $\theta$, which are based on the question and node embeddings,
    are used to sample a subgraph $z$ that is then used to predict the answer.
    }\label{fig:approach}
\end{figure}
We follow \citet{tilli} as our base system architecture, integrating the subset 
sampling methods described above to intrinsically sample a subgraph.
We replace the scene graph encoding modules based on GloVe~\cite{glove} embeddings to 
\gls*{clip}~\cite{clip} text token embeddings.
Additionally, we shift from percentage based top-$k$ sampling to fixed top-$k$ sampling to better control the size of the explanation subgraph.

The sampling process of our approach is illustrated in \cref{fig:approach}, where $\theta$ represents 
the prior scores, computed as a scaled dot-product between a representation of the question and the 
node embeddings of the graph.
The variable $z$ represents the sampled subgraph, where each entry indicates whether the 
corresponding node is included or excluded in the subgraph.
As a result, we obtain hard-attention masks, which are used to mask the graph's adjacency matrix and 
the node features.

\section{Experimental Results}

\paragraph{Setup}
We conduct experiments on the GQA dataset~\cite{gqa} using the ground-truth scene graphs for the 
\emph{training} and \emph{validation} splits.
All results are reported on the \emph{validation} split, as scene graphs for the \emph{testdev} split are not available.

\subsection{RQ1 -- Quantitative Results}
Due to the low question-answering performance of \gumbelst~($30.61\pm2.39$), we exclude it from the results.
The \gumbelst~estimator utilizes a relaxed approximation in its sampling mechanism, which introduces additional bias and variance, which might be the reason why it reduces the precision in selecting relevant nodes within the subgraph compared to the other methods.

The aggregated results of the experiments are summarized in \cref{tab:results}, 
while the table with individual runs and detailed results can be found in \cref{app:results}, 
specifically in \cref{app:tab:results}.
\begin{table}[htb!]
    \centering
    \resizebox{\linewidth}{!}{
    \begin{tabular}{lccc}
        \toprule
        Method & Accuracy & \textsc{At-coo} & \textsc{Qt-coo} \\
        \midrule
        \textsc{None} & 92.14$\pm$2.62 & -- & -- \\
        \midrule
        \aimle~& \textbf{93.34$\pm$0.99} & \textbf{92.66$\pm$3.23} & \textbf{80.86$\pm$6.84}  \\
        \simple~& 91.05$\pm$3.44 & 84.47$\pm$16.06 & 73.56$\pm$14.19 \\
        \imle~& 81.13$\pm$8.07 & 65.15$\pm$17.45 & 72.88$\pm$11.59 \\
        \bottomrule
    \end{tabular}
    }
    \caption{Mean and standard deviation for the performance metrics of each method. 
    The full table with the detailed results for each run can be found in~\cref{app:results}~\cref{app:tab:results}.}
    \label{tab:results}
\end{table}

We experimented with different top-$k$ values, batch sizes, and other hyperparameters to compare 
the performance of the models in terms of average answer accuracy, the \gls*{atcoo}, and the 
\gls*{qtcoo} (cf.~\cref{app:results:metrics} for a formal definition).
\aimle~achieved the highest answer accuracy across various top-$k$ values, exceeding $94\%$, 
closing the gap to the \textsc{None} baseline, where no subgraph sampling is applied 
(the alternative non-interpretable black-box approach).

\begin{figure}[!htb]
    \centering
    \includegraphics[width=1.0\linewidth]{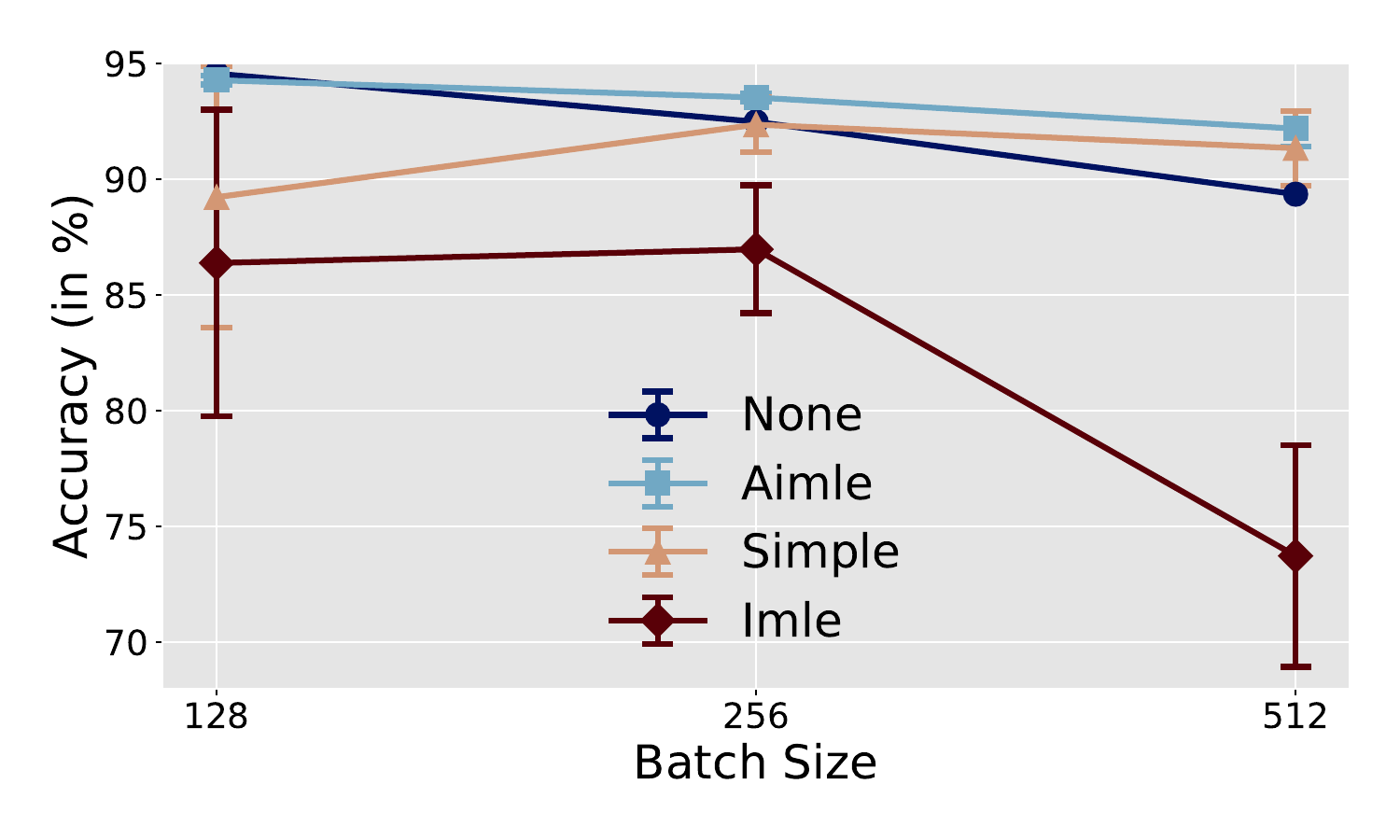}
    \caption{Model accuracy with respect to batch size.}
    \label{fig:acc_batch_sizes}
\end{figure}

\begin{figure}[!htb]
    \centering
    \includegraphics[width=1.0\linewidth]{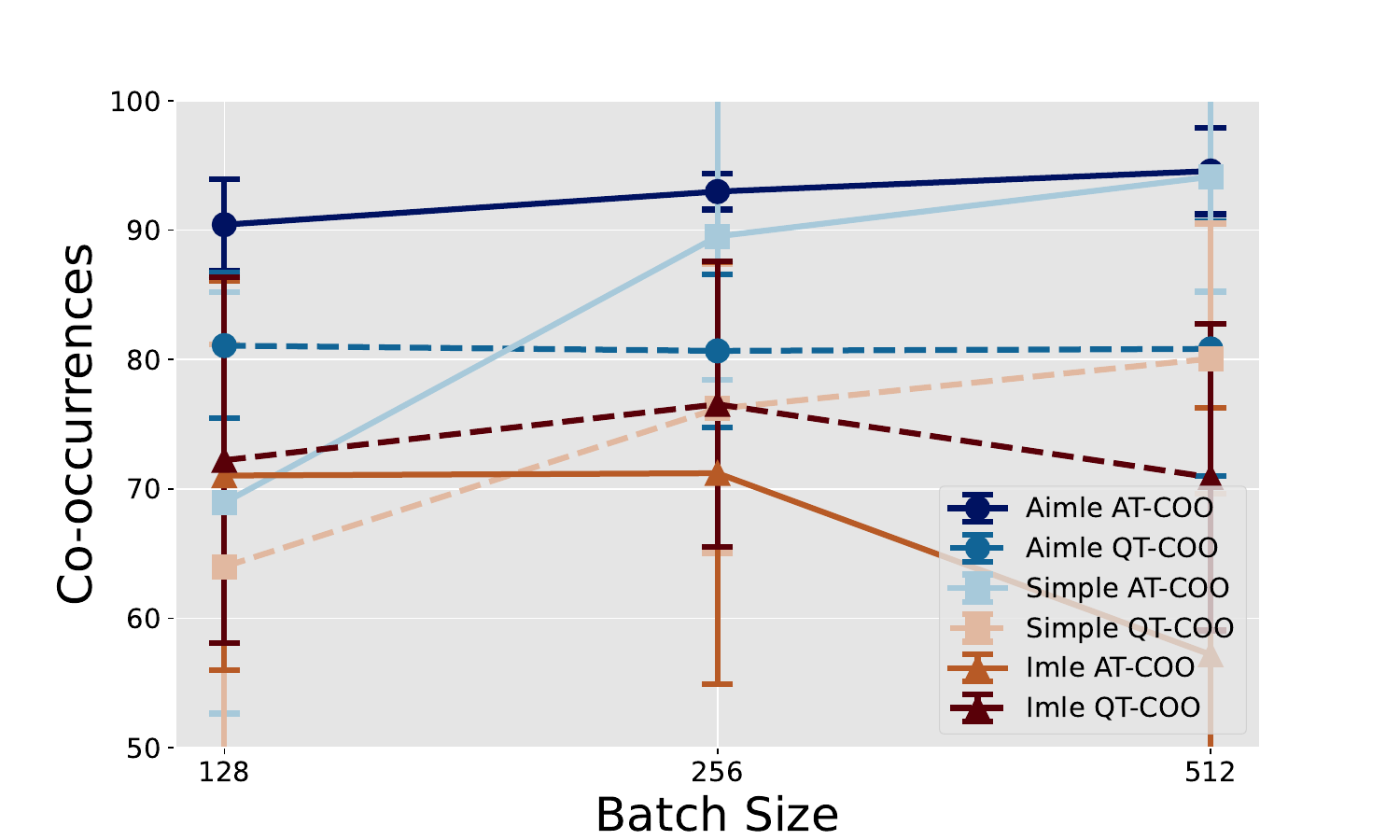}
    \caption{\gls*{atcoo} and \gls*{qtcoo} values with respect to batch size.}
    \label{fig:coo_batch_sizes}
\end{figure}

\paragraph{Effect of Batch Sizes}
While \cref{fig:acc_batch_sizes} shows a negative trend in model accuracy with increasing batch sizes, 
\cref{fig:coo_batch_sizes} indicates that the \gls*{atcoo} and \gls*{qtcoo} increase when training 
with larger batch sizes.
This suggests a trade-off between the accuracy of the model and the quality of the explanations, 
as larger batch sizes lead to better co-occurrences but worse accuracy.
\simple~underperformed compared to \aimle~with smaller batch sizes, suggesting that \simple~is more 
sensitive to this hyperparameter.
This effect is also reflected in the \gls*{atcoo} and \gls*{qtcoo} values, where \simple~achieved 
lower answer and question co-occurrences with smaller batch sizes.

We find that \imle~is an outlier in this regard, as it achieves higher co-occurrences with batch 
sizes of 128 and 256 than with 512.
This may be due to \imle's sensitivity to hyperparameters, particularly its method-specific 
hyperparameter $\lambda$, and the need for longer training to fully converge to the optimal solution.

\subsection{RQ2 -- Human Evaluation}

\paragraph{Design}
We recruited 60 participants through Prolific (\url{www.prolific.com}).
Each participant was presented with 18 pairwise comparisons between two explanatory subgraphs generated by 
two different subset sampling methods and asked to choose the preferred explanation.

\paragraph{Preference Estimation}
In total, we collected 1,080 pairwise comparisons, which were utilized to apply an extended 
Bradley-Terry model~\cite{exbtm} to estimate the relative preferences for the sampling methods.
The extended version introduces a parameter for ties, $\delta$, which adjusts for the likelihood 
of tied outcomes.

\paragraph{Results}
Due to all methods performing comparably well regarding the \gls*{atcoo} and \gls*{qtcoo}, we 
observe a large number of ties between methods, supported by the tie parameter $\delta=0.45$.
To mitigate the effect of ties in the Bradley-Terry model, we set the weight of ties to $\frac{1}{6}$.
\begin{table}[htb!]
    \centering
    \resizebox*{1.0\linewidth}{!}{
    \begin{tabular}{lcccc}
        \toprule
        Method & Favored & Ties & Unfavored & $\theta$\\
        \midrule
        \aimle~& 226 & 339 & 155 & 0.17 \\
        \simple~& 200 & 257 & 263 & -0.07 \\
        \imle~& 181 & 350 & 189 & -0.1 \\
        \bottomrule
    \end{tabular}
    }
    \caption{The individual effects parameters of the extended Bradley-Terry model.}
    \label{tab:individual-effects}
\end{table}
Thus, we emphasize the cases where the participants favored one method over another,
rather than focusing on ties where the preferred method was indistinguishable.
The individual effects parameters are shown in \cref{tab:individual-effects}.
According to the Bradley-Terry model ranking, \aimle~is favored over \simple~and \imle, while
\simple~is favored over \imle.

\paragraph{Ranking Consistency Analysis}
To verify whether the \gls*{atcoo} and \gls*{qtcoo} metrics generalize to comparisons between intrinsic 
subgraph sampling methods -- addressing \textbf{RQ2} -- we correlated the Bradley-Terry model's 
parameters with the \gls*{atcoo} and \gls*{qtcoo} metrics.
The results are shown in \cref{tab:correlation}.
\begin{table}[htb!]
    \centering
    \resizebox{\linewidth}{!}{
    \begin{tabular}{lccc}
        \toprule
        Metric & Pearson's $r$ & Spearman's $\rho$ & Kendall's $\tau$\\
        \midrule
        \textsc{At-coo} & 0.795 & 1.0 & 1.0 \\
        \textsc{Qt-coo} & 0.99 & 1.0 & 1.0 \\
        \bottomrule
    \end{tabular}
    }
    \caption{Correlation scores between the Bradley-Terry model's parameter $\theta$ and the \gls*{atcoo} and \gls*{qtcoo} metrics.}
    \label{tab:correlation}
\end{table}
We find that both metrics strongly correlate with $\theta$, suggesting the effectiveness of the 
\gls*{atcoo} and \gls*{qtcoo} metrics.
Consequently, the ranking of the intrinsic subgraph sampling methods according to the Bradley-Terry
model aligns with the ranking according to the \gls*{atcoo} and \gls*{qtcoo} metrics.

\section{Conclusion}\label{sec:conclusion}
We integrated and compared discrete subset sampling methods — \imle, \aimle, \simple, and \gumbelst — to intrinsically generate explanatory subgraphs in a \gls*{gvqa} system.
To answer \textbf{RQ1}, we found that \gumbelst~degraded performance to a non-competitive level, while the other methods achieved competitive results compared to the black-box approach.
Depending on the choice of hyperparameters, \aimle~and \simple~achieved the highest answer accuracies, 
\gls*{atcoo}, and \gls*{qtcoo} scores, while \imle~exhibited higher sensitivity to hyperparameter tuning.

In the human evaluation, that was conducted to answer \textbf{RQ2}, \aimle~was most favored by participants, followed by \simple~and \imle, 
indicating that \aimle~is the most promising method, requiring minimal tuning and performing well out-of-the-box.
For the other methods, more careful hyperparameter selection is necessary to achieve competitive results.

\section{Limitations}\label{sec:limitations}

The accuracy and reliability of the model are highly dependent on the quality and quantity of the 
training data. 
Biases, inconsistencies, or missing data can significantly affect the model's results and 
explanations. 
In our case, the provided ground-truth scene graphs contain errors or inaccuracies, 
which can negatively impact the system's performance and limit its applicability to real-world scenarios. 

The fixed top-$k$ sampling does not adapt to the complexity of individual questions. 
In real-world scenarios, the relevance of subgraph explanations could vary significantly, 
requiring a more dynamic top-$k$ selection to better tailor explanations to the complexity of the 
visual question. 
For some question types, the explanations might be too simplistic or overly complex, which may limit 
the interpretability.

Human evaluation is inherently subjective, which introduces variability in the assessment 
of model explanations.
Factors such as individual preferences, prior knowledge, personal biases, interpretation of explanations,
or experience with such systems can influence the evaluation results.

\section{Ethics Statement}\label{sec:ethics}

All participants provided informed consent prior to their involvement in the study. 
We offered a comprehensive explanation of the task and research goals and refrained from collecting 
any personally identifiable information from the users. 
All logs and survey responses were encrypted using an anonymous hash derived from the participant's
Prolific username. 
We confirmed the estimated time in our pilot study to ensure that the selected duration was below 
the median completion time.
The participants were compenstated for their time and effort with a wage that exceeded the minimum 
wage in the country of the study's origin based on the median completion time established through the 
pilot study.

\section{Acknowledgements}\label{sec:acknowledgements}
Funded by Deutsche Forschungsgemeinschaft (DFG, German Research Foundation) under Germany’s Excellence Strategy - EXC 2075 – 390740016. We acknowledge the support by the Stuttgart Center for Simulation Science (SimTech).

\bibliography{custom}

\begin{thebibliography}{30}
\providecommand{\natexlab}[1]{#1}

\bibitem[{Achiam et~al.(2023)Achiam, Adler, Agarwal, Ahmad, Akkaya, Aleman, Almeida, Altenschmidt, Altman, Anadkat et~al.}]{chatgpt}
Josh Achiam, Steven Adler, Sandhini Agarwal, Lama Ahmad, Ilge Akkaya, Florencia~Leoni Aleman, Diogo Almeida, Janko Altenschmidt, Sam Altman, Shyamal Anadkat, et~al. 2023.
\newblock Gpt-4 technical report.
\newblock \emph{arXiv preprint arXiv:2303.08774}.

\bibitem[{Agrawal et~al.(2018)Agrawal, Batra, Parikh, and Kembhavi}]{vqa_v2}
Aishwarya Agrawal, Dhruv Batra, Devi Parikh, and Aniruddha Kembhavi. 2018.
\newblock Don't just assume; look and answer: Overcoming priors for visual question answering.
\newblock In \emph{Proceedings of the IEEE conference on computer vision and pattern recognition}.

\bibitem[{Ahmed et~al.(2023)Ahmed, Zeng, Niepert, and Van~den Broeck}]{simple}
Kareem Ahmed, Zhe Zeng, Mathias Niepert, and Guy Van~den Broeck. 2023.
\newblock Simple: A gradient estimator for k-subset sampling.
\newblock In \emph{Proceedings of the 11th International Conference on Learning Representations}.

\bibitem[{Anil et~al.(2023)Anil, Borgeaud, Wu, Alayrac, Yu, Soricut, Schalkwyk, Dai, Hauth, Millican et~al.}]{gemini}
Rohan Anil, Sebastian Borgeaud, Yonghui Wu, Jean-Baptiste Alayrac, Jiahui Yu, Radu Soricut, Johan Schalkwyk, Andrew~M Dai, Anja Hauth, Katie Millican, et~al. 2023.
\newblock Gemini: A family of highly capable multimodal models.
\newblock \emph{arXiv preprint arXiv:2312.11805}.

\bibitem[{Ansel et~al.(2024)Ansel, Yang, He, Gimelshein, Jain, Voznesensky, Bao, Bell, Berard, Burovski, Chauhan, Chourdia, Constable, Desmaison, DeVito, Ellison, Feng, Gong, Gschwind, Hirsh, Huang, Kalambarkar, Kirsch, Lazos, Lezcano, Liang, Liang, Lu, Luk, Maher, Pan, Puhrsch, Reso, Saroufim, Siraichi, Suk, Zhang, Suo, Tillet, Zhao, Wang, Zhou, Zou, Wang, Mathews, Wen, Chanan, Wu, and Chintala}]{pytorch}
Jason Ansel, Edward Yang, Horace He, Natalia Gimelshein, Animesh Jain, Michael Voznesensky, Bin Bao, Peter Bell, David Berard, Evgeni Burovski, Geeta Chauhan, Anjali Chourdia, Will Constable, Alban Desmaison, Zachary DeVito, Elias Ellison, Will Feng, Jiong Gong, Michael Gschwind, Brian Hirsh, Sherlock Huang, Kshiteej Kalambarkar, Laurent Kirsch, Michael Lazos, Mario Lezcano, Yanbo Liang, Jason Liang, Yinghai Lu, C.~K. Luk, Bert Maher, Yunjie Pan, Christian Puhrsch, Matthias Reso, Mark Saroufim, Marcos~Yukio Siraichi, Helen Suk, Shunting Zhang, Michael Suo, Phil Tillet, Xu~Zhao, Eikan Wang, Keren Zhou, Richard Zou, Xiaodong Wang, Ajit Mathews, William Wen, Gregory Chanan, Peng Wu, and Soumith Chintala. 2024.
\newblock Pytorch 2: Faster machine learning through dynamic python bytecode transformation and graph compilation.
\newblock In \emph{Proceedings of the 29th ACM International Conference on Architectural Support for Programming Languages and Operating Systems, Volume 2}.

\bibitem[{Antol et~al.(2015)Antol, Agrawal, Lu, Mitchell, Batra, Zitnick, and Parikh}]{vqa}
Stanislaw Antol, Aishwarya Agrawal, Jiasen Lu, Margaret Mitchell, Dhruv Batra, C~Lawrence Zitnick, and Devi Parikh. 2015.
\newblock Vqa: Visual question answering.
\newblock In \emph{Proceedings of the IEEE international conference on computer vision}.

\bibitem[{Damodaran et~al.(2021)Damodaran, Chakravarthy, Kumar, Umapathy, Mitamura, Nakashima, Garcia, and Chu}]{understanding_sg_vqa}
Vinay Damodaran, Sharanya Chakravarthy, Akshay Kumar, Anjana Umapathy, Teruko Mitamura, Yuta Nakashima, Noa Garcia, and Chenhui Chu. 2021.
\newblock Understanding the role of scene graphs in visual question answering.
\newblock \emph{arXiv preprint arXiv:2101.05479}.

\bibitem[{Fey and Lenssen(2019)}]{pyg}
Matthias Fey and Jan~E. Lenssen. 2019.
\newblock Fast graph representation learning with {PyTorch Geometric}.
\newblock In \emph{ICLR Workshop on Representation Learning on Graphs and Manifolds}.

\bibitem[{Hildebrandt et~al.(2020)Hildebrandt, Li, Koner, Tresp, and G{\"u}nnemann}]{sg_vqa_survey}
Marcel Hildebrandt, Hang Li, Rajat Koner, Volker Tresp, and Stephan G{\"u}nnemann. 2020.
\newblock Scene graph reasoning for visual question answering.
\newblock \emph{arXiv preprint arXiv:2007.01072}.

\bibitem[{Hudson and Manning(2019)}]{gqa}
Drew~A Hudson and Christopher~D Manning. 2019.
\newblock Gqa: A new dataset for real-world visual reasoning and compositional question answering.
\newblock In \emph{Proceedings of the IEEE/CVF conference on computer vision and pattern recognition}.

\bibitem[{Hunter(2004)}]{exbtm}
David~R Hunter. 2004.
\newblock Mm algorithms for generalized bradley-terry models.
\newblock \emph{The annals of statistics}.

\bibitem[{Jang et~al.(2017)Jang, Gu, and Poole}]{gumbel-softmax-1}
Eric Jang, Shixiang Gu, and Ben Poole. 2017.
\newblock Categorical reparametrization with gumble-softmax.
\newblock In \emph{International Conference on Learning Representations}.

\bibitem[{Liang et~al.(2021)Liang, Jiang, and Liu}]{graphvqa}
Weixin Liang, Yanhao Jiang, and Zixuan Liu. 2021.
\newblock Graphvqa: Language-guided graph neural networks for graph-based visual question answering.
\newblock In \emph{Proceedings of the Third Workshop on Multimodal Artificial Intelligence}.

\bibitem[{Likert(1932)}]{likert}
Rensis Likert. 1932.
\newblock A technique for the measurement of attitudes.
\newblock \emph{Archives of psychology}.

\bibitem[{Liu et~al.(2023{\natexlab{a}})Liu, Li, Li, and Lee}]{improvedllava}
Haotian Liu, Chunyuan Li, Yuheng Li, and Yong~Jae Lee. 2023{\natexlab{a}}.
\newblock Improved baselines with visual instruction tuning.

\bibitem[{Liu et~al.(2024)Liu, Li, Li, Li, Zhang, Shen, and Lee}]{llavanext}
Haotian Liu, Chunyuan Li, Yuheng Li, Bo~Li, Yuanhan Zhang, Sheng Shen, and Yong~Jae Lee. 2024.
\newblock \href {https://llava-vl.github.io/blog/2024-01-30-llava-next/} {Llava-next: Improved reasoning, ocr, and world knowledge}.

\bibitem[{Liu et~al.(2023{\natexlab{b}})Liu, Li, Wu, and Lee}]{llava}
Haotian Liu, Chunyuan Li, Qingyang Wu, and Yong~Jae Lee. 2023{\natexlab{b}}.
\newblock Visual instruction tuning.

\bibitem[{Maddison et~al.(2017)Maddison, Mnih, and Teh}]{gumbel-softmax-2}
Chris~J Maddison, Andriy Mnih, and Yee~Whye Teh. 2017.
\newblock The concrete distribution: A continuous relaxation of discrete random variables.
\newblock In \emph{International Conference on Learning Representations}.

\bibitem[{Maddison et~al.(2014)Maddison, Tarlow, and Minka}]{gumbel-max}
Chris~J Maddison, Daniel Tarlow, and Tom Minka. 2014.
\newblock A* sampling.
\newblock \emph{Advances in neural information processing systems}.

\bibitem[{Mersha et~al.(2024)Mersha, Lam, Wood, AlShami, and Kalita}]{xai_survey}
Melkamu Mersha, Khang Lam, Joseph Wood, Ali AlShami, and Jugal Kalita. 2024.
\newblock Explainable artificial intelligence: A survey of needs, techniques, applications, and future direction.
\newblock \emph{Neurocomputing}.

\bibitem[{Minervini et~al.(2023)Minervini, Franceschi, and Niepert}]{aimle}
Pasquale Minervini, Luca Franceschi, and Mathias Niepert. 2023.
\newblock Adaptive perturbation-based gradient estimation for discrete latent variable models.
\newblock In \emph{Proceedings of the AAAI Conference on Artificial Intelligence}.

\bibitem[{Niepert et~al.(2021)Niepert, Minervini, and Franceschi}]{imle}
Mathias Niepert, Pasquale Minervini, and Luca Franceschi. 2021.
\newblock Implicit mle: backpropagating through discrete exponential family distributions.
\newblock \emph{Advances in Neural Information Processing Systems}.

\bibitem[{Papandreou and Yuille(2011)}]{perturb_map}
George Papandreou and Alan~L Yuille. 2011.
\newblock Perturb-and-map random fields: Using discrete optimization to learn and sample from energy models.
\newblock In \emph{2011 International Conference on Computer Vision}.

\bibitem[{Pennington et~al.(2014)Pennington, Socher, and Manning}]{glove}
Jeffrey Pennington, Richard Socher, and Christopher Manning. 2014.
\newblock {G}lo{V}e: Global vectors for word representation.
\newblock In \emph{Proceedings of the 2014 Conference on Empirical Methods in Natural Language Processing}.

\bibitem[{Qian et~al.(2024)Qian, Manolache, Ahmed, Zeng, Van~den Broeck, Niepert, and Morris}]{qianprobabilistically}
Chendi Qian, Andrei Manolache, Kareem Ahmed, Zhe Zeng, Guy Van~den Broeck, Mathias Niepert, and Christopher Morris. 2024.
\newblock Probabilistically rewired message-passing neural networks.
\newblock In \emph{International Conference on Learning Representations}.

\bibitem[{Radford et~al.(2021)Radford, Kim, Hallacy, Ramesh, Goh, Agarwal, Sastry, Askell, Mishkin, Clark et~al.}]{clip}
Alec Radford, Jong~Wook Kim, Chris Hallacy, Aditya Ramesh, Gabriel Goh, Sandhini Agarwal, Girish Sastry, Amanda Askell, Pamela Mishkin, Jack Clark, et~al. 2021.
\newblock Learning transferable visual models from natural language supervision.
\newblock In \emph{International conference on machine learning}.

\bibitem[{Tilli and Vu(2024)}]{tilli}
Pascal Tilli and Ngoc~Thang Vu. 2024.
\newblock Intrinsic subgraph generation for interpretable graph based visual question answering.
\newblock In \emph{Proceedings of the 2024 Joint International Conference on Computational Linguistics, Language Resources and Evaluation}.

\bibitem[{Touvron et~al.(2023)Touvron, Lavril, Izacard, Martinet, Lachaux, Lacroix, Rozi{\`e}re, Goyal, Hambro, Azhar et~al.}]{llama}
Hugo Touvron, Thibaut Lavril, Gautier Izacard, Xavier Martinet, Marie-Anne Lachaux, Timoth{\'e}e Lacroix, Baptiste Rozi{\`e}re, Naman Goyal, Eric Hambro, Faisal Azhar, et~al. 2023.
\newblock Llama: Open and efficient foundation language models.
\newblock \emph{arXiv preprint arXiv:2302.13971}.

\bibitem[{Wang et~al.(2023)Wang, Yasunaga, Ren, Wada, and Leskovec}]{vqa_gnn}
Yanan Wang, Michihiro Yasunaga, Hongyu Ren, Shinya Wada, and Jure Leskovec. 2023.
\newblock Vqa-gnn: Reasoning with multimodal knowledge via graph neural networks for visual question answering.
\newblock In \emph{Proceedings of the IEEE/CVF International Conference on Computer Vision}.

\bibitem[{Xie and Ermon(2019)}]{softsub-st}
Sang~Michael Xie and Stefano Ermon. 2019.
\newblock Reparameterizable subset sampling via continuous relaxations.
\newblock In \emph{Proceedings of the International Joint Conference on Artificial Intelligence}.

\end{thebibliography}

\appendix

\section{Background}\label{app:background}
\subsection{Visual Question Answering}
\gls{vqa} is a task in the field of \gls{ml} that involves a system's ability to answer questions 
about visual content, typically images or videos.
To generate an answer, the model is required to process the visual input as well as the 
natural language question in form of text.
For the visual modality, the model processes the images to implicitly or explicitly identify objects 
with their corresponding attributes, scene information, and relations among objects.
The most difficult aspect of the task is to combine these two information resources and learn to 
reason between the two modalities.
\gls{vqa} system can be applied in various applications, including assistive technologies or 
educational tools, offering insights into the interplay between language understanding and visual 
perception.

\subsection{Graph-based Visual Question Answering}
Structured representations of images as scene graphs make them particularly effective for answering 
questions that involve complex relationships or require an understanding of spatial or semantic 
interactions among multiple objects. 
The graph-based approach in \gls{gvqa} changes the reasoning process to identify paths or subgraphs 
that capture the information required to answer the question, which has the potential to enhance the 
system's interpretability.

\section{Experimental Results}\label{app:results}

\subsection{Metrics}\label{app:results:metrics}

\paragraph{Answer Token Co-Occurrences}
Let \( S = \{ s_1, \ldots, s_n \} \) be the set of subgraphs, where \( s_i = \{ v_{i1}, \ldots, v_{im} \} \) are node tokens for each subgraph \( s_i \). Let \( A = \{ a_1, \ldots, a_n \} \) be the set of answer tokens, and let \( a_i \in A \) be the specific answer token for each subgraph \( s_i \).
Let $\mathbb{I}(a, s)$ be an indicator function defined as
\begin{equation}
    \mathbb{I}(a, s) =
    \begin{cases}
    1 & \text{if } a \in s \\
    0 & \text{otherwise}
    \end{cases}
\end{equation}
Then the answer token co-occurrence can be computed as
\begin{equation}
    P_A = \frac{1}{n}\sum_{i=1}^{n} I(a_i, s_i)
\end{equation}

\paragraph{Question Token Co-Occurrences}
Let \( S = \{ s_1, \ldots, s_n \} \) be the set of subgraphs, where \( s_i = \{ v_{i1}, \ldots, v_{im} \} \) are node tokens for each subgraph \( s_i \). Let \( Q = \{ q_1, \ldots, q_k \} \) be the set of question tokens, and let \( Q_i \subseteq Q \) be the subset of question tokens related to the subgraph \( s_i \).
Define the indicator function \( \mathbb{I}(q, s) \) as
\begin{equation}
    \mathbb{I}(q, s) =
    \begin{cases}
    1 & \text{if } q \in s \\
    0 & \text{otherwise}
    \end{cases}
\end{equation}
Define the match ratio function \( R(Q_i, s) \) as
\begin{equation}
    R(Q_i, s) = \frac{\sum_{q \in Q_i} \mathbb{I}(q, s)}{|Q_i|}
\end{equation}
which computes the ratio of matching question tokens in \( Q_i \) that are present in the subgraph 
\( s \).
Then the question token co-occurrence can be computed as
\[
P_Q = \frac{1}{n} \sum_{i=1}^{n} R(Q_i, s_i)
\]
which computes the average match ratio across all subgraphs.

\subsection{Quantitative Results}
\begin{figure}[htb!]
    \centering
    \includegraphics[width=1.0\linewidth]{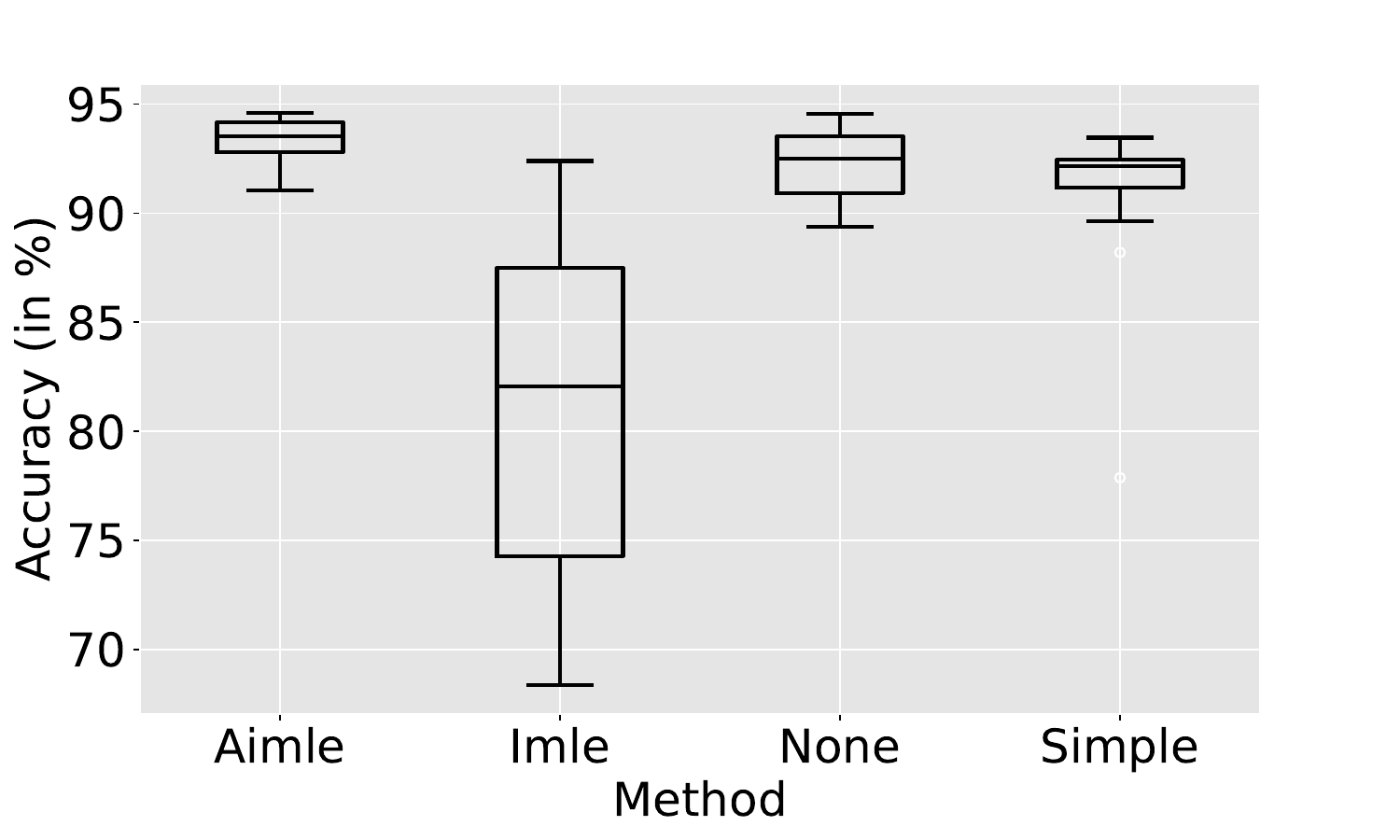}
    \caption{Accuracy per method across different top-$k$ values and batch sizes.}
    \label{app:fig:acc_boxplots}
\end{figure}

\paragraph{Boxplots}
\cref{app:fig:acc_boxplots} aggregates the results for each method across different top-$k$ values and batch sizes in form of boxplots.
The visualizations show the volatility of the accuracy of different models and indicate that \textsc{Aimle} and \textsc{Simple} are the most stable methods, while \textsc{Imle} requires more tuning of hyper-parameters to achieve similar results.

\begin{table*}[h!]
    \centering
    \begin{tabular}{lrccccccc}
        \toprule
        \textbf{Method} & Top-$k$ & Accuracy & Batch-Size & $N$-Epochs & \textsc{At-coo} & \textsc{Qt-coo} & $\lambda$ & $\tau$ \\
        \midrule

        \multirow{3}{*}{\textsc{None}}
        & -- & 94.57 & 128 & 50 & -- & -- & -- & -- \\
        & -- & 92.49 & 256 & 50 & -- & -- & -- & -- \\
        & -- & 89.36 & 512 & 50 & -- & -- & -- & -- \\
        
        \midrule
        
        \multirow{15}{*}{\textsc{Aimle}}
        & \multirow{3}{*}{2}
        & 94.26 & 128 & 50 & 90.67 & 76.31 & \textit{tuned} & 1 \\ 
        & & 93.45 & 256 & 50 & 92.62 & 71.06 & \textit{tuned} & 1 \\ 
        & & 92.90 & 512 & 50 & 96.90 & 66.49 & \textit{tuned} & 1 \\ 
        \cline{2-9}
        & \multirow{3}{*}{3}
        & 94.22 & 128 & 50 & 91.49 & 81.11 & \textit{tuned} & 1 \\ 
        & & 93.65 & 256 & 50 & 93.78 & 80.44 & \textit{tuned} & 1 \\ 
        & & 91.81 & 512 & 50 & 94.59 & 84.47 & \textit{tuned} & 1 \\ 
        \cline{2-9}
        & \multirow{3}{*}{4}
        & 94.17 & 128 & 50 & 84.55 & 75.02 & \textit{tuned} & 1 \\ 
        & & 93.74 & 256 & 50 & 93.30 & 86.38 & \textit{tuned} & 1 \\ 
        & & 91.05 & 512 & 50 & 89.13 & 76.25 & \textit{tuned}  & 1  \\
        \cline{2-9}
        & \multirow{3}{*}{5}
        & 94.61 & 128 & 30 & 94.10 & 84.45 & \textit{tuned} & 1 \\ 
        & & 93.51 & 256 & 50 & 90.81 & 80.85 & \textit{tuned} & 1 \\ 
        & & 92.52 & 512 & 50 & 94.59 & 84.46 & \textit{tuned} & 1 \\
        \cline{2-9}
        & \multirow{3}{*}{6}
        & 94.18 & 128 & 50 & 91.30 & 88.49 & \textit{tuned} & 1 \\ 
        & & 93.34 & 256 & 50 & 94.41 & 84.61 & \textit{tuned} & 1 \\
        & & 92.68 & 512 & 50 & 97.64 & 92.44 & \textit{tuned} & 1 \\

        \midrule
        
        \multirow{15}{*}{\textsc{Simple}}
        & \multirow{3}{*}{2}
        & 77.87 & 128 & 50 & 56.96 & 39.76 & -- & -- \\
        & & 91.04 & 256 & 50 & 66.91 & 56.42 & -- & -- \\
        & & 88.20 & 512 & 50 & 76.22 & 64.05 & -- & -- \\
        \cline{2-9}
        & \multirow{3}{*}{3}
        & 89.63 & 128 & 50 & 48.82 & 50.99 & -- & -- \\
        & & 92.97 & 256 & 50 & 83.68 & 67.09 & -- & -- \\
        & & 92.23 & 512 & 50 & 95.40 & 74.90 & -- & -- \\
        \cline{2-9}
        & \multirow{3}{*}{4}
        & 91.87 & 128 & 50 & 72.91 & 68.52 & -- & -- \\
        & & 93.46 & 256 & 50 & 92.28 & 74.15 & -- & -- \\
        & & 91.28 & 512 & 50 & 97.04 & 75.62 & -- & -- \\
        \cline{2-9}
        & \multirow{3}{*}{5}
        & 91.57 & 128 & 50 & 60.12 & 59.73 & -- & -- \\
        & & 93.05 & 256 & 50 & 93.42 & 81.19 & -- & -- \\
        & & 91.53 & 512 & 50 & 98.17 & 85.54 & -- & -- \\
        \cline{2-9}
        & \multirow{3}{*}{6}
        & 92.25 & 128 & 50 & 88.91 & 82.37 & -- & -- \\
        & & 90.52 & 256 & 50 & 94.54 & 82.52 & -- & -- \\
        & & 92.65 & 512 & 50 & 98.27 & 88.79 & -- & -- \\
        
        \midrule
        
        \multirow{12}{*}{\textsc{Imle}} 
        & \multirow{3}{*}{2}
        & 82.48 & 128 & 50 & 68.39 & 56.06 & 10 & 1 \\
        & & 83.38 & 256 & 50 & 72.74 & 61.78 & 10 & 1 \\
        & & 73.83 & 512 & 50 & 46.54 & 75.74 & 10 & 1 \\
        \cline{2-9}
        & \multirow{3}{*}{3}
        & 79.08 & 128 & 50 & 53.96 & 80.07 & 10 & 1 \\
        & & 87.74 & 256 & 50 & 47.67 & 88.47 & 10 & 1 \\
        & & 69.32 & 512 & 50 & 43.13 & 73.36 & 10 & 1 \\
        \cline{2-9}
        & \multirow{3}{*}{4}
        & 92.39 & 128 & 50 & 71.26 & 65.35 & 10 & 1 \\
        & & 90.03 & 256 & 50 & 82.25 & 78.75 & 10 & 1 \\
        & & 68.35 & 512 & 50 & 41.91 & 72.02 & 10 & 1 \\
        \cline{2-9}
        & \multirow{3}{*}{5}
        & 91.59 & 128 & 50 & 90.51 & 87.38 & 10 & 1 \\
        & & 86.77 & 256 & 50 & 82.18 & 77.15 & 10 & 1 \\
        & & 75.63 & 512 & 50 & 48.89 & 84.20 & 10 & 1 \\
        
        
        \bottomrule
    \end{tabular}
    \caption{Results of models trained with different top-$k$ subset sampling methods across different batch sizes.}
    \label{app:tab:results}
\end{table*}

\section{Human Evaluation}
We implemented a web application to perform a human evaluation.
The instructions for the participants can be found in \cref{app:human_eval:instr}.
The self-assessment of their prior knowledge is displayed in \cref{app:human_eval:demo}.

\subsection{Instructions}\label{app:human_eval:instr}
At the beginning of the web page, we provided the participants with the following instructions.

\emph{
The study contains 18 images for each participant. The image is not used as input to the model, it is only displayed as a reference. 
Next to the original image you can find the corresponding question the model answered.
The answer is displayed next to the prediction field. We also included the annotated ground-truth label for each question, i.e. the correct answer given by the dataset.
This states what should have been the correct answer according to the data. 
Some questions might be ambiguous unfortunately, but this should be not evaluated or considered in this study.
}

\emph{
Below the original image, you can find the corresponing graphs. 
The full graph (nodes colored in blue and green) is input to the model alongside the question.
The graph itself (all nodes colored in green and blue combined) might not be a perfect representation of the image. 
Nodes, which represent objects in the image, might be missing, or the annotation (the label/name) might be misleading.
We display the edges between nodes in the visualization of the graph, but we excluded the annotation (the name of the relation). Edges represent relations between objects, e.g. \textbf{a man holding a racket} would result in two nodes, \textbf{man} and \textbf{racket}, and one edge (relation) \textbf{holding} between them.
}

\emph{
We perform pair-wise comparisons between two explainability methods. Their explanations are displayed next to each other. 
All nodes colored in green are part of the subgraph that represents the explanation of model. 
All nodes colored in blue are excluded, so they are not part of the explanation. 
The explanatory subgraph (nodes in green) should support the predicted answer.
To judge which explantion you prefer, you should take the question and answer into account, and evaluate if the nodes in green form a more valid explanation than the other explanation.
}

\emph{
We only compare explanations that are of the same size, i.e. they contain equal number of nodes. Hence, we do not want to judge, if the explanatory subgraph consists of too many (green) nodes, but rather if the nodes included in the explanation capture the relevant information to answer the question. 
Please mark which of the explanations you prefer for each given pair.
}

\subsection{Compensation}
We estimated the reward for completing the task with a medium time of 15 min, above 
minimum wage with \pounds11.20/hr. 

\subsection{Demographics}\label{app:human_eval:demo}
We asked the participants to self assess their knowledge about \gls{ai} and \gls{xai} on a 
Likert~\cite{likert} scale from 1 to 5, where 1 corresponds to no knowledge and 5 to very good knowledge.
\begin{figure}[htb!]
    \centering
    \includegraphics[width=1.0\linewidth]{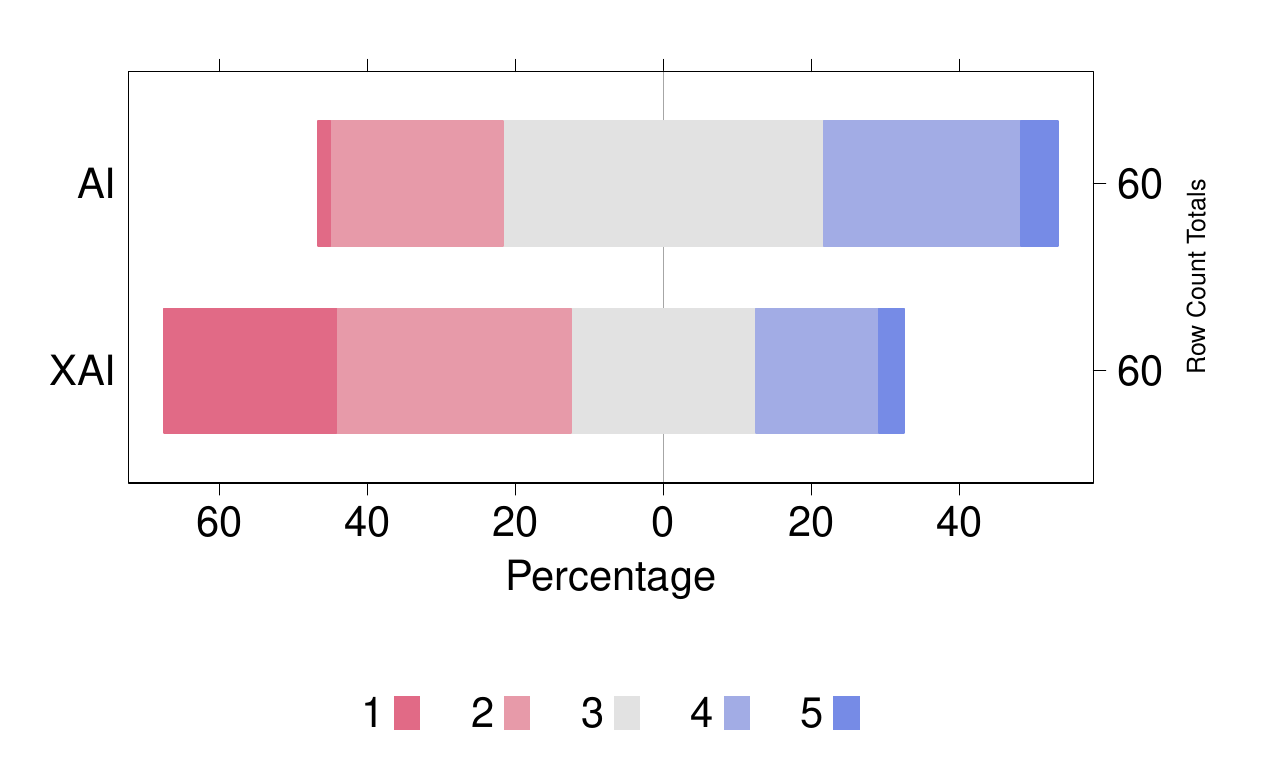}
    \caption{Likert scale responses to the questions about \gls{ai} and \gls{xai}.}
    \label{app:fig:likert}
\end{figure}
\cref{app:fig:likert} visualizes the distribution of the scores of the Likert scale responses.
While the majority of participants rated their knowledge about \gls{ai} and \gls{xai} as average,
the participants were generally more informed about \gls{ai} than \gls{xai}.

\begin{figure}[htb!]
    \centering
    \includegraphics[width=1.0\linewidth]{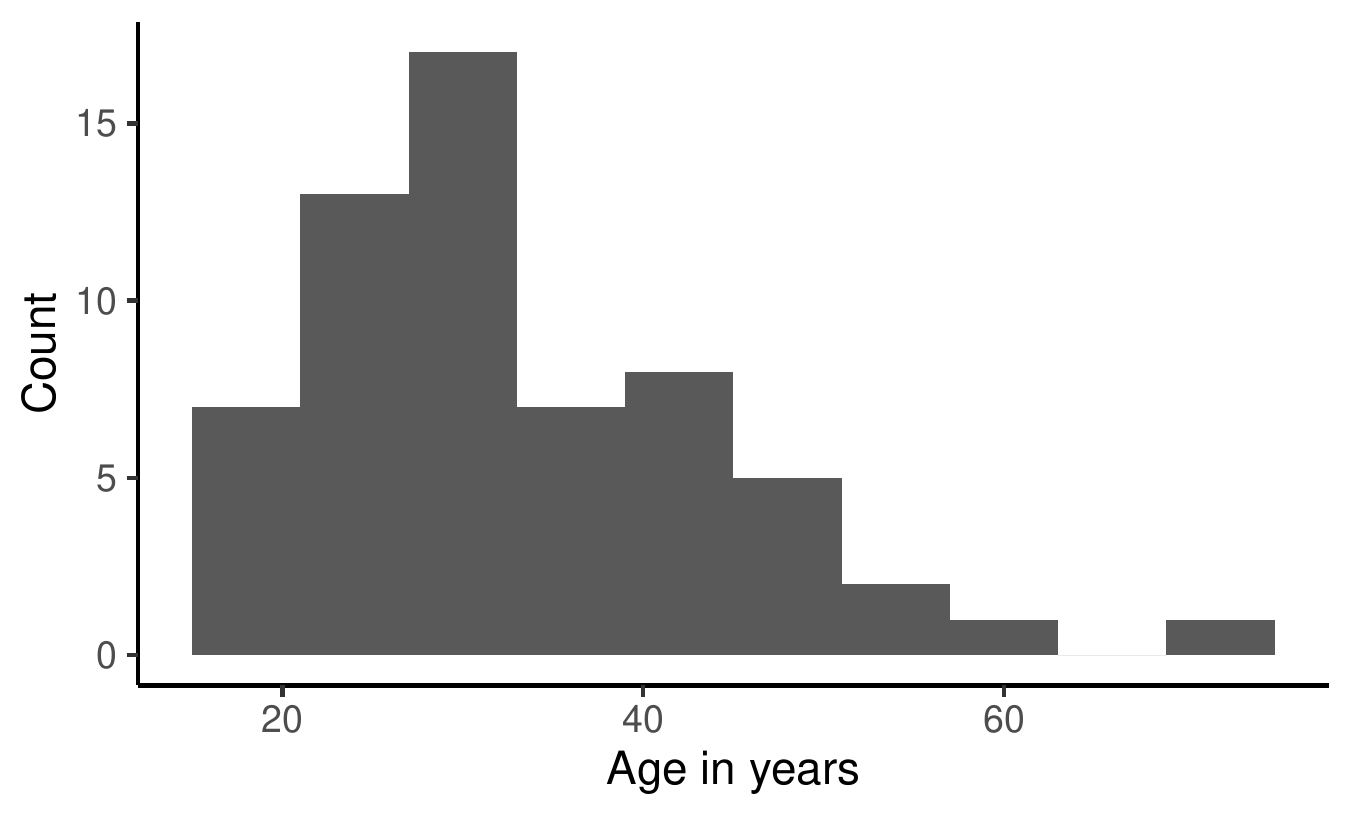}
    \caption{Age distribution of the participants.}
    \label{app:fig:age}
\end{figure}

The age distribution of the participants is shown in \cref{app:fig:age}.
The majority of participants were between 18 and 24 years old, with fewer participants over 35 years old.

\section{Implementation Details}\label{app:implementation}
We implemented our models using PyTorch 2~\cite{pytorch} and PyTorch Geometric~\cite{pyg}.
For the subset sampling methods, we use the official implementions of \imle~and \aimle~\cite{imle, aimle}, 
in combination with \simple~and \gumbelst~by~\citet{qianprobabilistically}.

\onecolumn

\section{Examples}
\begin{figure*}[h]
    \centering
    \includegraphics[width=1.0\linewidth]{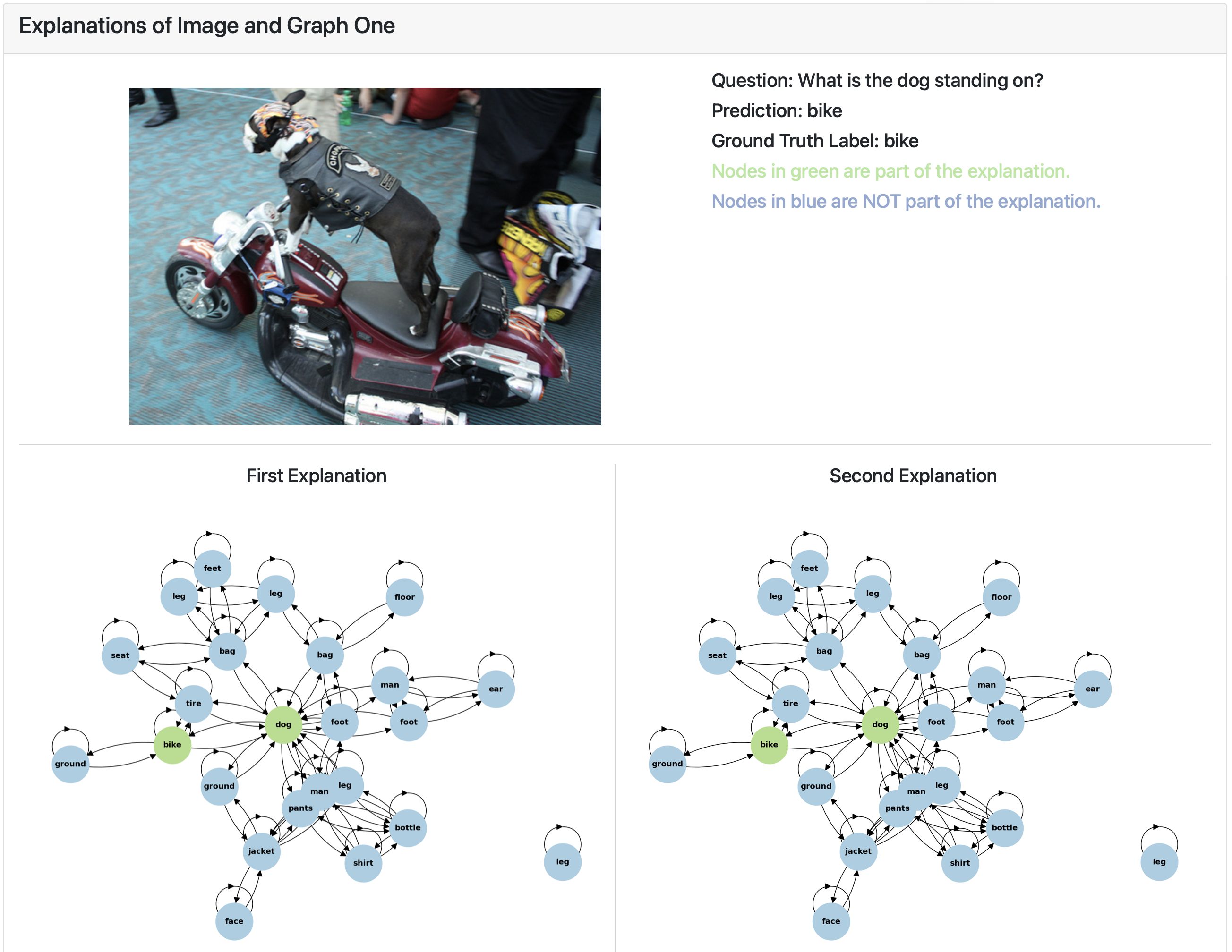}
    \caption{User interface for the human evaluation of two explanatory subgraphs, in green, and excluded nodes, in blue.}
    \label{app:fig:dog_bike}
\end{figure*}

\end{document}